\documentstyle[latex-acl,named]{article} 
\author{Walter Daelemans, Sabine Buchholz, Jorn Veenstra\\
	ILK, Tilburg University, PO-box 90153, NL 5000 LE Tilburg\\
	\tt{[walter,buchholz,veenstra@kub.nl]}}
\title{Memory-Based Shallow Parsing}

\begin{document} 

\maketitle

\begin{abstract}

We present a memory-based learning (MBL) approach to shallow parsing
in which POS tagging, chunking, and identification of syntactic
relations are formulated as memory-based modules. The experiments
reported in this paper show competitive results, the $F_{\beta=1}$ for
the Wall Street Journal (WSJ) treebank is: 93.8\% for NP chunking,
94.7\% for VP chunking, 77.1\% for subject detection and 79.0\% for
object detection.

\end{abstract}

\bibliographystyle{named}

\section{Introduction}
\label{introduction}

Recently, there has been an increased interest in approaches to
automatically learning to recognize shallow linguistic patterns in
text
\cite{Ramshaw+95,Vilain+96,Argamon+98,Buchholz98,Cardie+98,Veenstra98,Daelemans+99}.
Shallow parsing is an important component of most text analysis
systems in applications such as information extraction and summary
generation.  It includes discovering the
main constituents of sentences (NPs, VPs, PPs) and their heads, and
determining syntactic relationships like subject, object, adjunct
relations between verbs and heads of other constituents.

Memory-Based Learning (MBL) shares with other statistical and learning
techniques the advantages of avoiding the need for manual definition
of patterns (common practice is to use hand-crafted regular
expressions), and of being reusable for different corpora and
sublanguages.  The unique property of memory-based approaches which
sets them apart from other learning methods is the fact that they are
{\em lazy learners}: they keep all training data available for
extrapolation.  All other statistical and machine learning methods are
{\em eager\/} (or {\em greedy\/}) learners: They abstract knowledge structures or
probability distributions from the training data, forget the
individual training instances, and extrapolate from the induced
structures.  Lazy learning techniques have been shown to achieve
higher accuracy than eager methods for many language processing tasks.  A reason for this is the intricate interaction
between regularities, subregularities and exceptions in most language
data, and the related problem for learners of distinguishing noise
from exceptions.  Eager learning techniques abstract from what they
consider noise (hapaxes, low-frequency events, non-typical events)
whereas lazy learning techniques keep all data available, including
exceptions which may sometimes be productive.  For a detailed analysis
of this issue, see \cite{Daelemans+99}.  Moreover, the automatic
feature weighting in the similarity metric of a memory-based learner
makes the approach well-suited for domains with large numbers of
features from heterogeneous sources, as it embodies a
smoothing-by-similarity method when data is sparse \cite{Zavrel+97}.

In this paper, we will provide a empirical evaluation of the
MBL approach to syntactic analysis on a number of shallow pattern
learning tasks: NP chunking, VP chunking, and the assignment of
subject-verb and object-verb relations.  The approach is evaluated by
cross-validation on the WSJ treebank corpus \cite{Marcus+93}.
We compare the approach qualitatively and as far as possible
quantitatively with other approaches.

\section{Memory-Based Shallow Syntactic Analysis}
\label{MBL}

Memory-Based Learning (MBL) is a classification-based, supervised
learning approach: a memory-based learning algorithm constructs a
classifier for a task by storing a set of examples. Each example
associates a feature vector (the problem description) with one of a
finite number of classes (the solution). Given a new feature vector,
the classifier extrapolates its class from those of the most similar
feature vectors in memory. The metric defining similarity can be
automatically adapted to the task at hand.

In our approach to memory-based syntactic pattern recognition, we
carve up the syntactic analysis process into a number of such
classification tasks with input vectors representing a focus item and
a dynamically selected surrounding context.  As in Natural Language
Processing problems in general \cite{Daelemans95}, these
classification tasks can be segmentation tasks (e.g. decide whether a
focus word or tag is the start or end of an NP) or disambiguation
tasks (e.g. decide whether a chunk is the subject NP, the object NP or
neither). Output of some memory-based modules (e.g. a tagger or a
chunker) is used as input by other memory-based modules
(e.g. syntactic relation assignment).

Similar cascading ideas have been explored in other approaches to text
analysis: e.g. finite state partial parsing \cite{Abney96,Grefenstette96},
statistical decision tree parsing \cite{Magerman94}, maximum entropy
parsing \cite{Ratnaparkhi97}, and memory-based learning
\cite{Cardie94,Daelemans+96b}.

\subsection{Algorithms and Implementation}

For our experiments we have used TiMBL\footnote{TiMBL is available
from: http://ilk.kub.nl/}, an MBL software package developed in our
group~\cite{Daelemans+99b}. We used the following variants of MBL:

\begin{itemize} 
\item
{\sc ib1-ig}: The distance between a test item and each memory item is
defined as the number of features for which they have a different
value (overlap metric). Since in most cases not all features are
equally relevant for solving the task, the algorithm uses information
gain (an information-theoretic notion measuring the reduction of
uncertainty about the class to be predicted when knowing the value of
a feature) to weight the cost of a feature value mismatch during
comparison. Then the class of the most similar training item is predicted to be the class of the test item.
Classification speed is linear to the number of training instances times the number of features.
\item
{\sc IGTree}: {\sc ib1-ig} is expensive
in basic memory and processing requirements.  With {\sc IGTree}, an
oblivious decision tree is created with features as tests, and ordered
according to information gain of features, as a heuristic
approximation of the computationally more expensive pure MBL variants.
Classification speed is linear to the number of features times the average branching factor in the tree, which is less than or equal to the average number of values per feature.
\end{itemize}

For more references and information about these algorithms we refer to
\cite{Daelemans+99b,Daelemans+99}. In \cite{Daelemans+96b} both
algorithms are explained in detail in the context of MBT, a
memory-based POS tagger, which we presuppose as an available module in
this paper. In the remainder of this paper, we discuss results on the
different tasks in section {\em Experiments}, and compare our approach
to alternative learning methods in section {\em Discussion and Related Research}.

\section{Experiments}
\label{experiments}

We carried out two series of experiments. In the first we evaluated a
memory-based NP and VP chunker, in the second we used this chunker for
memory-based subject/object detection.

To evaluate the performance of our trained memory-based classifiers,
we will use four measures: accuracy (the percentage of correctly
predicted output classes), precision (the percentage of predicted
chunks or subject- or object-verb pairs that is correct), recall (the percentage of chunks or subject- or object-verb pairs to be predicted that is found), and $F_{\beta}$ \cite{VanRijsbergen79},
which is given by $\frac{(\beta^2+1).prec.rec}{\beta^2 . prec + rec}$,
with $\beta=1$. See below for an example.

For the chunking tasks, we evaluated the algorithms by cross-validation on all 
25 partitions of the WSJ treebank. Each partition in turn was selected
as a test set, and the algorithms trained on the remaining
partitions. Average precision and recall on the 25 partitions will be
reported for both the {\sc ib1-ig} and {\sc igtree} variants of MBL.
For the subject/object detection task, we used 10-fold cross-validation on treebank partitions 00--09.
In section {\em Related~Research} we will further evaluate our chunkers and
subject/object detectors.

\subsection{Chunking}
\label{chunking}

\begin{table*}[hbt]

\begin{center}
\begin{tabular}{|l|c|c|c|c|c|}
\hline Methods 	& context	& accuracy &	precision 	& recall 	& $F_{\beta=1}$	\\ 
\hline 
\hline
 & \multicolumn{5}{|c|}{NPs}\\
\hline
IGTree		& 2-1		& 97.5 & 91.8		& 93.1	& 92.4	\\ 
IB1-IG		& 2-1		& 98.0 & 93.7 		& 94.0 	& 93.8	\\
baseline words	& 0		& 92.9 & 76.2 		& 79.7 	& 77.9	\\
baseline POS	& 0		& 94.7 & 79.5 		& 82.4 	& 80.9	\\
\hline
 & \multicolumn{5}{|c|}{VPs}\\
\hline
IGTree		& 2-1		& 99.0  & 93.0		& 94.2	& 93.6	\\
IB1-IG		& 2-1		& 99.2 	& 94.0		& 95.5	& 94.7	\\
baseline words	& 0		& 95.5 	& 67.5 		& 73.4 	& 70.3	\\
baseline POS	& 0		& 97.3 	& 74.7 		& 87.7 	& 81.2	\\
\hline
\end{tabular}
\end{center}

\caption{Overview of the NP/VP chunking scores of 25-fold cross-validation on the WSJ using {\sc IB1-IG} with a
context of two words and POS right and one left, and of using {\sc IGTree}
with the same context. The baseline scores are computed with {\sc IGTree} using only the
focus POS tag or the focus word}
\label{ChunkingResults25cv}
\end{table*}

\begin{table*}[hbt]
\begin{small}
\begin{center}
\begin{tabular}{|l|r|r|r|ll|ll|ll|ll|ll|c|}
\hline
Feature & 1 & 2 & 3 & 4 & 5 & 6 & 7 & 8 & 9 & 10 & 11 & 12 & 13 & Class \\ \hline
Weight  & 39 & 40 & 4 & 3 & 2 & 10 & 12 & 18 & 29 & 18 & 31 & 13 & 24 & \\ \hline\hline
Inst.1  & -1 & 0 & 0 & seen & VBN &        - & - &       - & - &   sisters & PRP\$ &   seen & VBN  &  S \\ 
Inst.2  & 1 & 0 & 0 & seen & VBN  & sisters & PRP\$  & seen & VBN      & man & NN    & lately & RB     & O \\ 
Inst.3  &  2 & 0 & 0 & seen & VNB     & seen & VBN    & man & NN    & lately & RB         & . & .      & - \\ \hline
\end{tabular}
\caption{\label{instances}Some sample instances for the subject/object detection task. The second row shows the relative weight of the features (truncated and multiplied by 100; from one of the 10 cross-validation experiments). Thus the order of importance of the features is: 2, 1, 11, 9, 13, 10, 8, 12, 7, 6, 3, 4, 5. }
\end{center}
\end{small}
\end{table*}

Following \cite{Ramshaw+95} we defined chunking as a tagging task,
each word in a sentence is assigned a tag which indicates whether this
word is inside or outside a chunk. We used as tagset:
\begin{description}
\item[I\_NP] inside a baseNP.
\item[O] outside a baseNP or a baseVP.
\item[B\_NP] inside a baseNP, but the preceding word is in another baseNP.
\item[I\_VP and B\_VP] are used in a similar fashion.
\end{description}
Since baseNPs and baseVPs are non-overlapping and non-recursive these
five tags suffice to unambiguously chunk a sentence.
For example, the sentence:
\begin{quote}
\begin{small}
[$_{NP}$ Pierre Vinken $_{NP}$] , [$_{NP}$ 61 years $_{NP}$] old ,
[$_{VP}$ will join $_{VP}$] [$_{NP}$ the board $_{NP}$] as [$_{NP}$ a
nonexecutive director $_{NP}$] [$_{NP}$ Nov. 29 $_{NP}$] .
\end{small}
\end{quote}
should be tagged as:

\begin{quote}
\begin{small}
Pierre$_{I\_NP}$ Vinken$_{I\_NP}$ ,$_O$ 61$_{I\_NP}$ years$_{I\_NP}$ old$_O$ ,$_O$
will$_{I\_VP}$ join$_{I\_VP}$ the$_{I\_NP}$ board$_{I\_NP}$ as$_O$ a$_{I\_NP}$
nonexecutive$_{I\_NP}$ director$_{I\_NP}$ Nov.$_{B\_NP}$ 29$_{I\_NP}$ .$_O$
\end{small}
\end{quote}

Suppose that our classifier erroneously tagged {\em director\/} as $B\_NP$ instead of $I\_NP$, but classified the rest correctly. Accuracy would then be $\frac{17}{18}=0.94$. The resulting chunks would be {\em [$_{NP}$ a
nonexecutive $_{NP}$] [$_{NP}$  director $_{NP}$]} instead of {\em [$_{NP}$ a nonexecutive director $_{NP}$]} (the other chunks being the same as above). Then out of the seven predicted chunks, five are correct (precision$=\frac{5}{7}=71.4\%$) and from the six chunks that were to be found, five were indeed found (recall$=\frac{5}{6}=83.3\%$). $F_{\beta=1}$ is 76.9\%.

The features for the experiments are the word form and
the POS tag (as provided by the WSJ treebank) of the two words to the left, the focus word, and one
word to the right. For the results see Table~\ref{ChunkingResults25cv}.

The baseline for these experiments is computed with {\sc IB1-IG}, with as
only feature: i) the focus word, and ii) the focus POS tag.

The results of the chunking experiments show that accurate chunking is
possible, with $F_{\beta=1}$ values around 94\%.

\subsection{Subject/Object Detection}

Finding a subject or object (or any other relation of a
constituent to a verb) is defined in our classification-based approach
as a mapping from a pair of words (the verb and the head of the
constituent) and a representation of its context to a class describing
the type of relation (e.g. subject, object, or neither).  A verb can
have a subject or object relation to more than one word in case of NP
coordination, and a word can be the subject of more than one verb in
case of VP coordination.

\subsection{Data Format}
\begin{table*}[hbt]
\centerline{
\begin{tabular}{|l||r|r|r|r||r|r|r||r|r|r|}
\hline
& \multicolumn{4}{|c||}{Together} & \multicolumn{3}{|c||}{Subjects} & \multicolumn{3}{|c|}{Objects} \\ \hline
\# relations & \multicolumn{4}{|c||}{51629} & \multicolumn{3}{|c||}{32755} & \multicolumn{3}{|c|}{18874} \\ \hline
Method & acc. & prec. & rec. & $F_{\beta=1}$ & prec. & rec. & $F_{\beta=1}$ & prec. & rec. & $F_{\beta=1}$ \\ \hline\hline
Random baseline            & &  3.9 &  3.9 &  3.9 &  4.5 &  4.5 &  4.5 &  2.7 &  2.5 &  2.6 \\
Heuristic baseline         & & 65.9 & 66.5 & 66.2 & 69.3 & 61.6 & 65.2 & 61.6 & 75.1 & 67.7 \\\hline
IGTree                     & 96.9 & 79.5       & 73.2       & 76.2       & 80.9       & 71.4       & 75.8       & 77.2       & 76.4 & 76.8 \\
IB1-IG                     & 96.6 & 74.4       & {\bf 76.9} & 75.6       & 76.2       & {\bf 76.9} & 76.5       & 71.5       & {\bf 76.7} & 74.0 \\ \hline
IGTree \& IB1-IG unanimous & 97.4 & {\bf 89.8} & 68.6       & {\bf 77.8} & {\bf 89.7} & 67.6       & {\bf 77.1} & {\bf 89.8} & 70.4 & {\bf 79.0} \\ \hline
\end{tabular}
}
\caption{\label{results_so}Results of the 10-fold cross validation experiment on the subject--verb/object--verb relations data. We trained one classifier to detect subjects as well as objects. Its performance can be found in the column {\em Together}. For expository reasons, we also mention how well this classifier performs when computing precision and recall for subjects and objects separately.} 
\end{table*}

In our representation, the tagged and chunked sentence
\begin{quote}
\label{s_o_example} {\bf [NP} {\em
My\/}/PRP\$ {\em sisters\/}/NNS {\bf NP] [VP} {\em have\/}/VBP {\em
not\/}/RB {\em seen\/}/VBN {\bf VP] [NP} {\em the\/}/DT {\em old\/}/JJ
{\em man\/}/NN {\bf NP]} {\em lately\/}/RB ./.  
\end{quote}
will result in the instances in Table~\ref{instances}.

Classes are {\em S(ubject)\/}, {\em O(bject)\/} or ``-'' (for anything else).  Features are: 
\begin{description}
\item[1] the distance from the verb to
the head (a chunk just counts for one word; a negative distance means
that the head is to the left of the verb), 
\item[2] the number of
other baseVPs between the verb and the head (in the current setting,
this can maximally be one), 
\item[3] the number of commas between the
verb and the head, 
\item[4] the verb, and 
\item[5] its POS tag, 
\item[6--9] the two left context words/chunks of the head, represented by
the word and its POS 
\item[10--11] the head itself, and 
\item[12--13] its right context word/chunk.  
\end{description}
Features one to three are numeric features. This property can only be 
exploited by {\sc IB1-IG}. {\sc IGTree} treats them as symbolic. We also tried four additional features that indicate the sort of chunk (NP, VP or none) of the head and the three context elements respectively. These features did not improve performance, presumably because this information is mostly inferrable from the POS tag.

To find subjects and objects in a test sentence, the sentence is first
POS tagged (with the Memory-Based Tagger MBT) and chunked (see section {\em Experiments: Chunking}). Subsequently, all chunks are reduced to their
heads.\footnote{By definition, the head is the rightmost word of a
baseNP or baseVP.} 

Then an instance is constructed for every pair of a baseVP and another word/chunk head provided they are not too distant from each other in the sentence. A crucial point here is the definition of ``not too distant''. If our definition is too strict, we might exclude too many actual subject-verb or object-verb pairs, which will result in low recall. If the definition is too broad, we will get very large training and test sets. This slows down learning and might even have a negative effect on precision because the learner is confronted with too much ``noise''. Note further that defining distance purely as the number of intervening words or chunks is not fully satisfactory as this does not take clause structure into account. As one clause normally contains one baseVP, we developped the idea of counting intervening baseVPs. Counts on the treebank showed that less than 1\% of the subjects and objects are separated from their verbs by more than one other baseVP. We therefore construct an instance for every pair of a baseVP and another word/chunk head if they have not more than one other baseVP in between them.\footnote{The following sentence shows a subject-verb pair (in bold) with one intervening baseVP (in italics): \newline
[$_{NP}$~The {\bf plant}~$_{NP}$], [$_{NP}$~which~$_{NP}$] [$_{VP}$~{\it is owned\/}~$_{VP}$] by [$_{NP}$~Hollingsworth \& Vose Co.~$_{NP}$] , [$_{VP}$~{\bf was}~$_{VP}$] under [$_{NP}$~contract~$_{NP}$] with [$_{NP}$~Lorillard~$_{NP}$] [$_{VP}$~to make~$_{VP}$] [$_{NP}$~the cigarette filters~$_{NP}$] . \newline   
The next example illustrates the same for an object-verb pair:\newline   
Along [$_{NP}$~the way~$_{NP}$] , [$_{NP}$~he~$_{NP}$] [$_{VP}$~{\bf meets}~$_{VP}$] [$_{NP}$~a solicitous Christian chauffeur~$_{NP}$] [$_{NP}$~who~$_{NP}$] [$_{VP}$~{\it offers\/}~$_{VP}$] [$_{NP}$~the hero~$_{NP}$] [$_{NP}$~God~$_{NP}$] [$_{NP}$~'s phone number~$_{NP}$] ; and [$_{NP}$~the Sheep {\bf Man}~$_{NP}$] , [$_{NP}$~a sweet, roughhewn figure~$_{NP}$] [$_{NP}$~who~$_{NP}$] [$_{VP}$~wears~$_{VP}$] -- [$_{NP}$~what else~$_{NP}$] -- [$_{NP}$~a sheepskin~$_{NP}$] .}

These instances are classified by
the memory-based learner.  For the training material, the POS tags and
chunks from the treebank are used directly. Also, subject-verb and
object-verb relations are extracted to yield the class values.

\subsubsection{Results and discussion}
\begin{table*}[hbt]

\begin{center}
\begin{tabular}{|l|l|l|l|l|l|l|}
\hline 
Method 		& Tagger	& accuracy & precision 	& recall 	& $F_{\beta=1}$	\\ 
\hline 
\hline
A,D\&K 		& Brill 	& -- 	& 91.6 	& 91.6 	& 91.6 \\ 
R\&M 		& Brill 	& 97.4 	& 92.3 	& 91.8 	& 92.0 \\ 
C\&P 		& Brill		& -- 	& 90.7 	& 91.1 	& 90.9 \\ 
\hline 
IB1-IG	 	& Brill 	& 97.2 & 91.5 	& 91.3 	& 91.4	\\ 
IB1-IG	 	& MBT	 	& 97.3 & 91.6 	& 91.5 	& 91.6	\\ 
IB1-IG	 	& WSJ 		& 97.6 & 92.2 	& 92.5 	& 92.3 \\ 
IB1-IG,POSonly 	& WSJ 		& 96.9 & 90.3 	& 90.1 	& 90.2	\\ 
\hline

\end{tabular}
\end{center}
\caption{Comparison of MBL and MBSL on same dataset of several
classifiers, the experiments with {\sc IB1-IG} are all carried out with a
context of five words and POS left and three right}
\label{israel-comp-np}
\end{table*}

The results in Table \ref{results_so} show that finding (unrestricted)
subjects and objects is a hard task. The baseline of classifying
instances at random (using only the probability distribution of the
classes) is about 4\%. Using the simple heuristic of classifying each
(pro)noun directly in front of resp.\ after the verb as {\em S\/}
resp.\ {\em O\/} yields a much higher baseline of about
66\%. Obviously, these are the easy cases. {\sc IGTree}, which is the better
overall MBL algorithm on this task, scores 10\% above this baseline,
i.e.\ 76.2\%. The difference in accuracy between {\sc IGTree} and {\sc IB1-IG} is only 0.3\%. In terms of F-values, {\sc IB1-IG} is better for finding
subjects, whereas {\sc IGTree} is better for objects. We also note that
{\sc IGTree} always yields a higher precision than recall, whereas {\sc IB1-IG}
does the opposite. 

{\sc IGTree} is thus more ``cautious'' than {\sc IB1-IG}. Presumably, this is due to the word-valued features. Many test instances contain a word not occurring in the training instances (in that feature). In that case, search in the {\sc IGTree} is stopped and the default class for that node is used. As the ``-'' class is more than ten times more frequent than the other two classes, there is a high chance that this default is indeed the ``-'' class, which is always the ``cautious'' choice. {\sc IB1-IG}, on the other hand, will not stop on encountering an unseen word, but will go on comparing the rest of the features, which might still opt for a non-``-'' class. The differences in precision and recall surely are a topic for further research. So far, this observation led us to combine both algorithms
by classifying an instance as {\em S\/} resp. {\em
O\/} only if both algorithms agreed and as ``-'' otherwise. The
combination yields higher precision at the cost of recall,
but the overall effect is certainly positive ($F_{\beta=1}=77.8\%$).

\section{Discussion and Related Research}
\label{related}

In \cite{Argamon+98}, an alternative approach to
memory-based learning of shallow patterns, memory-based sequence
learning (MBSL), is proposed. In this approach, tasks such as base NP
chunking and subject detection are formulated as separate
bracketing tasks, with as input the POS tags of a sentence. For every input
sentence, all possible bracketings in context (situated contexts) are
hypothesised and the highest scoring ones are used for generating a
bracketed output sentence. The score of a situated hypothesis depends
on the scores of the tiles which are part of it and the degree to
which they cover the hypothesis. A tile is defined as a substring of
the situated hypothesis containing a bracket, and the score of a tile
depends on the number of times it is found in the training material
divided by the total number of times the string of tags occurs
(i.e. including occurrences with another or no bracket).  The approach
is memory-based because all training data is kept available. Similar
algorithms have been proposed for grapheme-to-phoneme conversion by
\cite{Dedina+91}, and \cite{Yvon96}, and the approach could be seen as
a linear algorithmic simplification of the DOP memory-based approach
for full parsing \cite{Bod95}.  In the remainder of this section, we
show that an empirical comparison of our computationally simpler MBL
approach to MBSL on their data for NP chunking, subject, and object
detection reveals comparable accuracies.

\subsection{Chunking}
\begin{table*}[hbt]
\centerline{
\begin{tabular}{|l||r|r|r||r|r|r|}
\hline
& \multicolumn{3}{|c||}{Subjects} & \multicolumn{3}{|c|}{Objects}\\ \hline
\# subsequences & \multicolumn{3}{|c||}{3044} & \multicolumn{3}{|c|}{1626}\\ \hline
Method & prec. & rec. & $F_{\beta=1}$ & prec. & rec. & $F_{\beta=1}$ \\ \hline\hline
A,D\&K                        & 88.6 & 84.5 & {\bf 86.5} & 77.1 & 89.8 & 83.0 \\ \hline
IGTree                        & 79.9 & 71.7 & 75.6 & 84.4 & 85.8 & 85.1 \\
IB1-IG                        & 84.7 & 81.6 & 83.1 & 87.3 & 85.8 & {\bf 86.5} \\ \hline
IB1-IG POS only               & 83.5 & 77.9 & 80.6 & 76.1 & 83.3 & 79.6 \\
IB1-IG without chunks         & 29.2 & 24.4 & 26.6 & 85.0 & 18.5 & 30.4 \\
IB1-IG with treebank chunks   & 89.4 & 88.6 & 89.0 & 91.9 & 91.3 & 91.6 \\ \hline
\end{tabular}
}
\caption{Comparison of MBL and MBSL on subject/object detection
as formulated by Argamon et al.}
\label{israel-comp-so}
\end{table*}

For NP chunking, {\cite{Argamon+98} used data extracted from section
15-18 of the WSJ as a fixed train set and section 20 as a fixed test
set, the same data as \cite{Ramshaw+95}.  To find the optimal setting
of learning algorithms and feature construction we used 10-fold cross
validation on section 15; we found {\sc IB1-IG} with a context of five words
and POS-tags to the left and three to the right as a good parameter
setting for the chunking task; we used this setting as the default
setting for our experiments. For an overview of the results see
Table~\ref{israel-comp-np}.  Since part of the chunking errors could
be caused by POS errors, we also compared the same baseNP chunker on
the same corpus tagged with i) the Brill tagger as used in
\cite{Ramshaw+95}, ii) the Memory-Based Tagger (MBT) as described
in \cite{Daelemans+96b}. We also present the results of
\cite{Argamon+98}, \cite{Ramshaw+95} and \cite{Cardie+98} in
Table~\ref{israel-comp-np}. The latter two use a transformation-based
error-driven learning method \cite{Brill92}. In \cite{Ramshaw+95},
the method is used for NP chunking, and in \cite{Cardie+98} the
approach is indirectly used to evaluate corpus-extracted NP chunking
rules.

As \cite{Argamon+98} used only POS information for their MBSL
chunker, we also experimented with that option (POSonly in the
Table). Results show that adding words as information provides useful
information for MBL (see Table~\ref{israel-comp-np}).

\subsection{Subject/object detection}

For subject/object detection, we trained our algorithm on section
01--09 of the WSJ and tested on Argamon et al.'s test data (section
00).  We also used the treebank POS tags instead of MBT. For
comparability, we performed two separate learning experiments. The verb windows
are defined as reaching only to the left (up to one intervening
baseVP) in the subject experiment and only to the right (with no
intervening baseVP) in the object experiment. The relational output of
MBL is converted to the sequence format used by MBSL. The
conversion program first selects one relation in case of coordinated
or nested relations. For objects, the actual conversion is trivial:
The V--O sequence extends from the verb up to the head ({\em seen the
old man} for the example sentence on page \pageref{s_o_example}). 
In the case of subjects, the
S--V sequence extends from the beginning of the baseNP of the head up
to the first non-modal verb in the baseVP ({\em My sisters
have}). The program also uses filters to model some restrictions of
the patterns that Argamon et al.\ used for data extraction. They
extracted e.g.\ only objects that immediately follow the verb.

The results in Table~\ref{israel-comp-so} show that highly comparable
results can be obtained with MBL on the (impoverished) definition of
the subject-object task.
{\sc IB1-IG} as well as {\sc IGTree} are better than
MBSL on the object data. They are however worse on the subject
data. Two factors may have influenced this result. Firstly, more than
17\% of the precision errors of {\sc IB1-IG} concern cases in which the word
proposed by the algorithm is indeed the subject according to the
treebank, but the corresponding sequence is not included in
Argamon et al.'s test data due to their restricted extraction
patterns. Secondly, there are cases for which MBL correctly found the
head of the subject, but the conversion results in an incorrect
sequence. These are sentences like {\em ``All [NP the man NP] [NP 's
friends NP] came.''\/} in which {\em all\/} is part of the subject
while not being part of any baseNP.

Apart from using a different algorithm, the MBL experiments also
exploit more information in the training data than MBSL does. Ignoring
lexical information in chunking and subject/object detection decreased
the $F_{\beta=1}$ value by 2.5\% for subjects and 6.9\% for
objects. The bigger influence for objects may be due to verbs that
take a predicative object instead of a direct one. Knowing the lexical
form of the verb helps to make this distinction. In addition, time
expressions like ``(it rained) last week'' can be distinguished from
direct objects on the basis of the head noun.  Not chunking the text
before trying to find subjects and objects decreases F-values by more
than 50\%. Using the ``perfect'' chunks of the treebank, on the other
hand, increases F by 5.9\% for subjects and 5.1\% for objects. These
figures show how crucial the chunking step is for the succes of our
method.

\subsection{General}

Clear advantages of MBL are its efficiency (especially when using
{\sc IGTree}), the ease with which information apart from POS tags can be
added to the input (e.g. word information, morphological information,
wordnet tags, chunk information for subject and object detection), and
the fact that NP and VP chunking and different types of relation
tagging can be achieved in one classification pass. It is unclear how
MBSL could be extended to incorporate other sources of information
apart from POS tags, and what the effect would be on performance.
More limitations of MBSL are that it cannot find nested sequences,
which nevertheless occur frequently in tasks such as subject
identification\footnote{e.g.\ [SV John, who [SV I like SV], is SV] angry.}, and that it does not mark heads.

\section{Conclusion}

We have developed and empirically tested a memory-based learning (MBL)
approach to shallow parsing in which POS tagging, chunking, and
identification of syntactic relations are formulated as memory-based
modules. A learning approach to shallow parsing allows for fast
development of modules with high coverage, robustness, and
adaptability to different sublanguages.  The memory-based algorithms
we used ({\sc IB1-IG} and {\sc IGTree}) are simple and efficient supervised
learning algorithms. Our approach was evaluated on NP and VP chunking, and subject/object
detection (using output from the chunker). $F_{\beta=1}$ scores are
93.8\% for NP chunking, 94.7\% for VP chunking, 77.1\% for subject
detection and 79.0\% for object detection. The accuracy and efficiency
of the approach are encouraging (no optimisation or post-processing of
any kind was used yet), and comparable to or better than
state-of-the-art alternative learning methods.

We also extensively compared our approach to a recently proposed new
memory-based learning algorithm, memory-based sequence learning (MBSL,
\cite{Argamon+98} and showed that MBL, which is a computationally
simpler algorithm than MBSL, is able to reach similar precision and
recall when restricted to the MBSL definition of the NP chunking,
subject detection and object detection tasks. More importantly, MBL is
more flexible in the definition of the shallow parsing tasks: it
allows nested relations to be detected; it allows the addition and
integration into the task of various additional sources of information
apart from POS tags; it can segment a tagged sentence into different
types of constituent chunks in one pass; it can scan a chunked
sentence for different relation types in one pass (though separating subject-verb detection from object-verb detection is surely an option that must be investigated). 

In current research we are extending the approach to other types of
constituent chunks and other types of syntactic relations. Combined
with previous results on PP-attachment \cite{Zavrel+97b}, the results presented here will be
integrated into a complete shallow parser.

\section*{Acknowledgements}
This research was carried out in the context of the ``Induction of
Linguistic Knowledge'' (ILK) research programme, supported partially
by the Foundation of Language, Speech and Knowledge (TSL), which is
funded by the Netherlands Organisation for Scientific Research
(NWO). The authors would like to thank the other members of the ILK
group for the fruitful discussions and comments.

\end{document}